\documentclass[nai,crcready]{iosart2x}

\usepackage{xspace}

\newcommand{\EL}{\ensuremath{\mathcal{E\!L}}\xspace}

\newcommand{\onto}[1]{\ensuremath{\mathsf{#1}}}

\usepackage{amsmath,amsthm}
\usepackage{booktabs}
\usepackage{subfig}

\usepackage[colorinlistoftodos,prependcaption,textsize=tiny]{todonotes}

\newif\ifproofread

\pubyear{0000}
\volume{0}
\firstpage{1}
\lastpage{1}

\proofreadtrue

\begin{document}
\begin{frontmatter}

\title{On the Multiple Roles of Ontologies in Explainable AI}
\runtitle{On the Multiple Roles of Ontologies in Explainable AI}


\begin{aug}
\author[A]{\inits{R.}\fnms{Roberto} \snm{Confalonieri}\ead[label=e1]{roberto.confalonieri@unipd.it}%
\thanks{Corresponding author. \printead{e1}.}}
\author[B]{\inits{G.}\fnms{Giancarlo} \snm{Guizzardi}\ead[label=e2]{g.guizzardi@utwente.nl}}
\address[A]{Department of Mathematics `Tullio-Levi Civita', \orgname{University of Padua},
Padova, \cny{Italy}\printead[presep={\\}]{e1}}
\address[B]{Department of Semantics, Cybersecurity \& Services (SCS), \orgname{University of Twente},
Enschede, \cny{The Netherlands}\printead[presep={\\}]{e2}}
\end{aug}


\begin{abstract}
This paper discusses the different roles that explicit knowledge, in particular ontologies, can play in Explainable AI and in the development of human-centric explainable systems and intelligible explanations. We consider three main perspectives in which ontologies can contribute  significantly, namely reference modelling, common-sense reasoning, and knowledge refinement and complexity management. We overview some of the existing approaches in the literature, and we position them according to these three proposed perspectives. The paper concludes by discussing what challenges still need to be addressed to enable ontology-based approaches to explanation and to evaluate their human-understandability and effectiveness. 
\end{abstract}

\begin{keyword}
\kwd{Explainable AI}
\kwd{Intelligible explanations}
\kwd{Applied Ontologies}
\end{keyword}

\end{frontmatter}




\section{Introduction}

Explainable AI (XAI) has emerged as a critical aspect in the development of transparent and trustworthy Artificial Intelligence (AI) systems, as highlighted in recent research~\cite{IF-2023}. The importance of incorporating explanation capabilities in intelligent systems extends beyond considerations of user rights and acceptance. Providing explainability is crucial for designers and developers to improve system robustness, enable diagnostics for bias prevention, address fairness and discrimination concerns, and foster trust among users regarding the decision-making process~\cite{Ribeiro2016,FUZZ-IEEE-2021}. Moreover, explicability is considered one of the essential ethical dimensions for autonomous systems \cite{guizzardi2023ontology}.

The concept of interpretability in AI systems has been recognised since the 1990s~\cite{WIREs-2020}. However, it has recently gained significant attention and became a prominent research focus in computer science and other disciplines due to the proliferation of big data and the introduction of various data protection regulations in the development of AI systems. For instance, the General Data Protection Regulation (GDPR) mandates that individuals have the legal right to receive explanations for decisions made by algorithms that may impact them, as stated in Article 22 of~\cite{GDPR2016}. This policy highlights the pressing importance of transparency and interpretability in the design of AI systems and algorithms. 

XAI focuses on developing new approaches for explanations of black box models by trying to achieve good explainability without sacrificing system performance~\cite{IF-2023}. One typical approach is the extraction of local and global post-hoc explanations that approximate the behaviour of a black-model model by means of an interpretable proxy. Other approaches are based on hybrid or neuro-symbolic AI systems, advocating a tight integration between symbolic and non-symbolic knowledge structures, e.g., by combining symbolic and statistical methods of reasoning~\cite{neurai3ed2022}.  By combining the strengths of statistical analysis, which provides insights into the internal workings of models, with the power of symbolic knowledge represented by ontologies, researchers aim to achieve a more comprehensive and robust approach to explainability. 

The construction of truly explainable systems is widely seen as one of the grand challenges facing AI today~\cite{Kautz_2022}. However, there is no consensus regarding how to achieve this, with proposed techniques in the literature ranging from inductive logic programming~\cite{10092808}, logical tensor networks~\cite{LogicTensorNet2022},  Markov logic networks~\cite{MarkovLogicNetworks2016} and logical neural networks~\cite{logical-nn-2020}, to name a few.   
What it seems widely accepted is that knowledge representation---in its many incarnations---is a key asset to enact hybrid systems, and it can pave the way towards the creation of transparent and human-centred explainable knowledge-enabled systems. 

Explainable knowledge-enabled systems are a class of systems defined as~\cite{Chari2020b}  
``AI systems that include a representation of domain knowledge in the field of application, have mechanisms to incorporate the context of users, are interpretable, and host explanation facilities that generate user-comprehensible, context-aware, and provenance-enabled explanations of the mechanistic functioning of the AI system and the knowledge used.'' 
This class of systems will 
exhibit a series of properties among which modularity, interpretability, support for provenance, adaptation to user's needs, explanation facilities, integration and access to knowledge, and compliance with standards and obligations. A similar but less demanding definition for explainable systems is proposed by~\cite{Doran2017}, where 
automated reasoning is central to output crafted explanations without requiring human post processing as final step of the generative process.

This paper explores the role of explicit knowledge representation artifacts (symbolic structures), such as ontologies and knowledge graphs, in Explainable AI, particularly regarding trustworthy decision support systems and the generation of human-understandable explanations. Firstly, we provide a concise introduction to ontologies and their typical conceptualisation and formalisation in computer science. Next, we present and elaborate on three perspectives that demonstrate how formal knowledge can contribute to the development of intelligible decision systems and explanations. In support of these perspectives, we summarise existing works that align with each viewpoint. Throughout the article we provide several examples that illustrate the main concepts. Finally, the paper outlines several future challenges associated with the advancement of human-centric explainable systems and intelligible explanations.

\section{Ontologies and their Role in XAI}\label{sec:ontologies}

Several recent surveys and position papers~\cite{IF-2023,WIREs-2020,Mittelstadt2019ExplainingAI,Chari2020c} emphasise the importance of designing explainability solutions that cater to diverse purposes and stakeholders, and highlight the limitations of existing approaches in supporting comprehensive human-centric Explainable AI. Current methods predominantly concentrate on specific types and formats of explanations, primarily focusing on the mechanistic aspects that explain {\em how} decisions are made. However, they often overlook the need to address the more fundamental question of {\em why} decisions are made, such as offering causal and counterfactual explanations~\cite{Miller2017}.

In addition, it is important to acknowledge that explanations rely on background knowledge. This background knowledge encompasses both the decision being explained and the intended recipient of the explanation. Explainability techniques must bridge the communication gap between the AI system and the users of the explanations, tailoring their outputs to different user groups. Consequently, if the explanations are not easily understandable by the users, they may be compelled to seek additional knowledge to obtain reliable insights and avoid drawing false conclusions.

The use of explicit knowledge, such as ontologies, knowledge graphs, or other forms of structured formal knowledge, can potentially help to bridge these gaps~\cite{TIDDI2022103627}. There has been a renewed interest in incorporating explicit knowledge into AI in recent years. Ontologies and knowledge graphs have been successfully applied in various domains, including knowledge-aware news recommender systems~\cite{iana2022survey}, semantic data mining and knowledge discovery~\cite{RistoskiP16,dbpedia-swj}, as well as natural language understanding~\cite{MSGraph12,conceptnet2017}. These applications highlight the value of leveraging explicit knowledge to enhance AI systems and address the challenges of explainability and user comprehension. 

In the subsequent sections, we will commence by offering a concise introduction to ontologies, as they are commonly understood in AI, in particular, in areas such as knowledge representation and the Semantic Web. Subsequently, we will explore the potential role that ontologies can assume in the context of Explainable AI. 

\subsection{Ontologies}

Within the realm of computer science, ontologies serve as a formal method for representing the structure of a specific domain. They capture the essential entities and relationships that arise from observing and understanding the domain, enabling its comprehensive description~\cite{GuarinoOS09}. 

To illustrate the concept further, let us consider a simple conceptualisation of the domain of the university and its employees. In an ontology, the entities within this domain can be organised into concepts and relations using unary and non-unary (typically, binary) predicates. At the core of the ontology there is a hierarchy of concepts, known as a taxonomy. For instance, if our focus is on academic roles, we might have relevant concepts such as {\em Person}, {\em Researcher} and {\em Professor}. In this case, {\em Person} would be a super-concept of {\em Researcher}  and {\em Professor}. Additionally, a relevant relation could be {\em CollaboratesWith} which captures the connections between individuals. Within the ontology, a specific researcher employed by the university would be an instance of the corresponding concept, linking them to a broader domain structure.

From a formal representation perspective, an ontology can be defined as a collection of axioms formulated using a suitable logical language. These axioms serve to express the intended semantics of the ontology, providing a conceptual framework for understanding a specific domain. It is worth noting that axioms play a crucial role in constraining the interpretations of the language used to formalise the ontology. They define the intended models that correspond to the conceptualisation of the domain while excluding unintended interpretations. In this way, axioms help establish the boundaries and constraints of the ontology, ensuring its coherence and consistency. As an example, we can formulate simple axioms to define the properties of relations within our ontology. We can state that the relation {\em SupervisedBy} is asymmetric and intransitive, meaning that a junior researcher can be supervised by a professor, but not the other way around. On the other hand, the relation {\em CollaboratesWith} is symmetric, irreflexive, and non-transitive. These axioms help define the intended semantics of these relations within the domain. Several formal languages for representing ontologies exist, varying from schema languages (e.g., RDF schema) to more expressive logics (e.g., First Order and Higher-order logics, Modal Logic, Description Logics), to representation languages that go beyond formal semantics and make an explicit commitment to domain-independent foundational categories (e.g., OntoUML \cite{guizzardi2022ufo}). Clearly, expressive logical languages capture richer intended semantics, but do often not allow for sound and complete reasoning and if they do, reasoning sometimes remains untractable. 

\begin{figure}[!t]
\resizebox{\columnwidth}{!}{
\fbox{
	\begin{minipage}[t]{12cm}
            \footnotesize
		\begin{tabular}{ll}
		    {\em TBox:} \\
			\onto{Person \sqsubseteq \top}  \qquad & , \qquad  \onto{Title \sqsubseteq \top} \\
			
			\onto{Researcher \sqsubseteq Person \sqcap \exists hasTitle.PhD}  \qquad & , \qquad \onto{PhD \sqsubseteq Title}  \\
			
			\onto{Professor \sqsubseteq Researcher \sqcap \exists hasTitle.Tenured } \qquad & , \qquad  \onto{Tenured  \sqsubseteq Title}  \\
			{\em ABox:} \\
			\onto{Researcher(Bob)} \qquad   & , \qquad \onto{Professor(Mike)}  \\
            
            \onto{CollaboratesWith(Bob,Mike)} \qquad   & , \qquad \onto{SupervisedBy(Bob,Mike)}
		\end{tabular}
\end{minipage}
}}
\caption{An ontology excerpt for the university domain formalised in DL. $\sqsubseteq$ is the subsumption relation, $\sqcap$ is conjunction, and $\exists$ is the existential relation. $\top$ is the top concept in the ontology. The TBox axioms state that `every researcher is a person with a PhD title', and `every professor is a researcher with a tenured position'. The Abox axioms state that Bob is a researcher, Mike is a professor, Bob and Mike collaborate, and Bob is supervised by Mike.}
\label{fig:ontology}
\end{figure}

Description Logics (DL)~\cite{BaaderDLH03} are one of the most well-known knowledge representation languages used to model ontologies in AI. They are of particular interest because they were created with the focus on tractable reasoning, and they provide the underpinning semantics of the W3C Web Ontology Language (OWL). Given a set of concept names, a set of role names, and some connectives that vary on the DL used, a DL ontology consists of two sets of axioms: the so-called TBox (terminological box) and the ABox (assertional box). In general, the TBox contains axioms describing relations between concepts, while the ABox contains axioms stating individuals (instances). For example, the statement `every researcher is a person with a PhD title' belongs to the TBox, while `Bob is a researcher' belongs to the ABox. Figure~\ref{fig:ontology} shows a DL specification that we will consider here as an ontology for our simple conceptualisation of the university domain \EL~\cite{DBLP:conf/ijcai/BaaderBL05}.\footnote{\EL is a lightweight DL allowing 
only for conjunctions and existential restrictions. It supports polynomial reasoning and is widely used to specify biomedical ontologies, see e.g.,~\cite{bioportal}.} Notice that the example shows a very simple formalisation of the university domain; different formalisations may indeed exist, with different levels of details.


Ontologies played a crucial role as enabling technologies in the development of the Semantic Web~\cite{bernerslee2001semantic}. The Semantic Web aims to annotate data on the web with semantic information, enabling computers to interpret and process data effectively. Although the Semantic Web did not fully realise all of its envisioned potential, ontologies have regained popularity in recent years, partly due to the re-emergence of knowledge graphs. Notably, the introduction of Google's Knowledge Graph in 2012 contributed significantly to this resurgence.\footnote{\url{https://www.blog.google/products/search/introducing-knowledge-graph-things-not/}, last accessed on 2023/07/31.} Knowledge graphs, powered by ontologies, have proven to be valuable tools for organising and structuring vast amounts of data, enabling efficient data retrieval, knowledge discovery, and semantic reasoning. Reasoning over ontologies or knowledge graphs can be performed by means of standard knowledge representation formalisms (e.g., RDF, RDF Schema, OWL) and query languages (e.g., SPARQL, Cypher, Gremlin) just to name a few of them. 

In recent years, a number of knowledge graphs have become available on the Web, offering valuable structured information. One prominent example is DBpedia~\cite{dbpedia-swj}, which constructs a knowledge graph by automatically extracting key-value pairs from Wikipedia infoboxes. These pairs are then mapped to the DBpedia ontology using crowdsourcing efforts. Another notable knowledge graph is ConceptNet~\cite{conceptnet2017}, a freely available linguistic knowledge graph that integrates information from various sources such as WordNet, Wikipedia, DBpedia, and OpenCyc. These knowledge graphs provide valuable resources for semantic understanding, information retrieval, and knowledge discovery. For a more comprehensive description of knowledge graphs and related standards, we recommend referring to the works cited in~\cite{TIDDI2022103627,kg-book-2021}. 

\subsection{Ontologies and XAI}

The knowledge representation language used to formalise an ontology provides support for both standard and non-standard reasoning tasks. Standard reasoning tasks involve checking various properties, such as subsumption and satisfaction. Concept subsumption determines if the description of one concept, for example, {\em Researcher}, is more general than the description of another concept, like {\em Professor}. Concept satisfaction, on the other hand, determines if a concept can be instantiated by an individual. Other standard reasoning tasks, such as query answering, are not semantically distinct from subsumption and satisfaction. Non-standard reasoning tasks have emerged to address new challenges in knowledge-based systems. These tasks include concept abduction and explanation, concept similarity, concept rewriting, and unification, among others. These non-standard reasoning tasks sometimes extend beyond the traditional subsumption and satisfaction checks, offering more advanced capabilities for knowledge-based systems.


Given this background, we believe that ontologies can play a crucial role in the realm of Explainable AI. In particular, their adoption can significantly contribute to the development of explainable knowledge-enabled systems from various perspectives: 

\begin{itemize}
\item {\bf Reference Modelling}: Ontologies provide sound and consensual models. By utilising ontologies (as reference models), it becomes possible to capture system requirements effectively and promote the reuse of components. Additionally, ontologies contribute to system accountability and facilitate knowledge sharing and reuse.

\item {\bf Common-Sense Reasoning}: Ontologies serve as a foundation for enriching explanations with context-aware semantic information. They support common-sense reasoning, which is essential for effectively transmitting knowledge to users. This capability enhances the comprehensibility of explanations for human understanding.

\item {\bf Knowledge Refinement and Complexity Management}: Ontologies enable the representation of abstraction and refinement, fundamental mechanisms underlying human reasoning. These mechanisms provide opportunities for integrating knowledge from diverse sources and customising the specificity and generality levels of explanations based on specific user profiles or target audiences. Moreover, systematic approaches grounded in Ontology (capital `O', i.e., as a philosophical discipline) can be used to reveal domain notions that are fundamental for explaining the propositional knowledge contained in an ontology.
\end{itemize}

By leveraging ontologies in these ways, the development of explainable knowledge-enabled systems can be enhanced, ensuring a solid conceptual basis, facilitating understanding through common-sense reasoning, and enabling flexible knowledge abstraction and refinement.

In the following sections, we overview some of the existing approaches in the literature, and we position them according to these three proposed perspectives. 

\section{Reference Modelling}

Ontologies serve as explicit symbolic models that facilitate the conceptualisation and rationalisation of information systems. By utilising ontologies, it is possible to explicit represent the various components within knowledge-based systems (KBSs), including their interactions (orchestration) and information workflows (choreography). The goal of ontologies in this context was to promote knowledge-based interoperability and component reusability, ultimately leading to increased systems transparency.

In KBSs, domain ontologies were created to represent the elements of a shared domain conceptualisation, and task ontologies
were created to represent generic problem-solving methods and facilitate the reuse of task-dependent knowledge across diverse domains and applications~\cite{guarino1998formal}. 
In this context, the Unified Problem-Solving Method description Language (UPML) served as a relevant resource for representing tasks and problem-solving methods as reusable, domain-independent components~\cite{Fensel2003}. Another example is the Web Service Modeling Ontology (WSMO), which focused on describing different aspects associated with Semantic Web Services~\cite{Fensel2002}. These ontology resources played crucial roles in facilitating the representation and standardisation of knowledge in their respective classes of applications.  


In the context of XAI, ontologies can play a crucial role as a common reference model for specifying explainable systems, i.e., as a model that addresses the area of explainable systems itself as a domain whose shared conceptualisation needs to be articulated and explicitly represented. Several studies have explored this avenue (e.g.,~\cite{BarredoArrieta2020,Nunes2017ASystems,Wang2019,Tiddi2015,Chari2023}) highlighting the necessity of a shared interchange model for addressing the factors involved in explainable systems. To achieve this, they proposed taxonomies and ontologies to model key notions of the XAI domain including: explanations, users, the mapping of end-user requirements to specific explanation types, as well as to the AI capabilities of systems.  

Nunes and Jannach~\cite{Nunes2017ASystems}  conducted a systematic literature review that examined the characteristics of explanations provided to users. Their work focused on aspects such as content, presentation, generation, and evaluation of explanations. They proposed a taxonomy that encompasses various explanation goals and different forms of knowledge that comprise the explanation components. 

Arrieta et al.~\cite{BarredoArrieta2020} present a taxonomy that established a mapping between deep learning models and the explanations they generate. Furthermore, they identified the specific features within these models that are responsible for generating these explanations. Their research contributes to the understanding of the relationship between model characteristics and the interpretability of their outputs. Their taxonomy covers different types of explanations that are produced by sub-symbolic models, including {\em simplification}, {\em explanation by examples}, {\em local explanations}, {\em text explanations}, {\em visual explanations}, {\em feature relevance}, and {\em explanations by transparent models}. 

In the study by Wang et al.~\cite{Wang2019}, the authors introduced a conceptual framework that elucidates how human reasoning processes can be integrated with explainable AI techniques. This framework establishes connections between different facets of explainability, such as explanation goals and types, as well as human reasoning mechanisms and AI methods. Notably, it facilitates a deeper understanding of the parallels between human reasoning and the generation of explanations by AI systems. By leveraging this conceptual framework, researchers and practitioners can gain insights into the interplay between human cognition and explainable AI.

Tiddi et al.~\cite{Tiddi2015} proposed an ontology design pattern specifically tailored for explanations. The authors observed that while the components of explanations may vary across different fields, there exist certain atomic components that can represent generic explanations. These atomic components include the associated {\em event}, the underlying {\em theory}, the {\em situation} to which the explanations are applied, and the {\em condition} that the explanations rely on. Additionally, the authors employ standard nomenclature such as {\em explanandum} (referring to what is being explained) and {\em explanans}  (referring to what does the explaining) to further clarify the structure of explanations.

\begin{figure}[!t]
    \centering
    \includegraphics[width=0.65\textwidth]{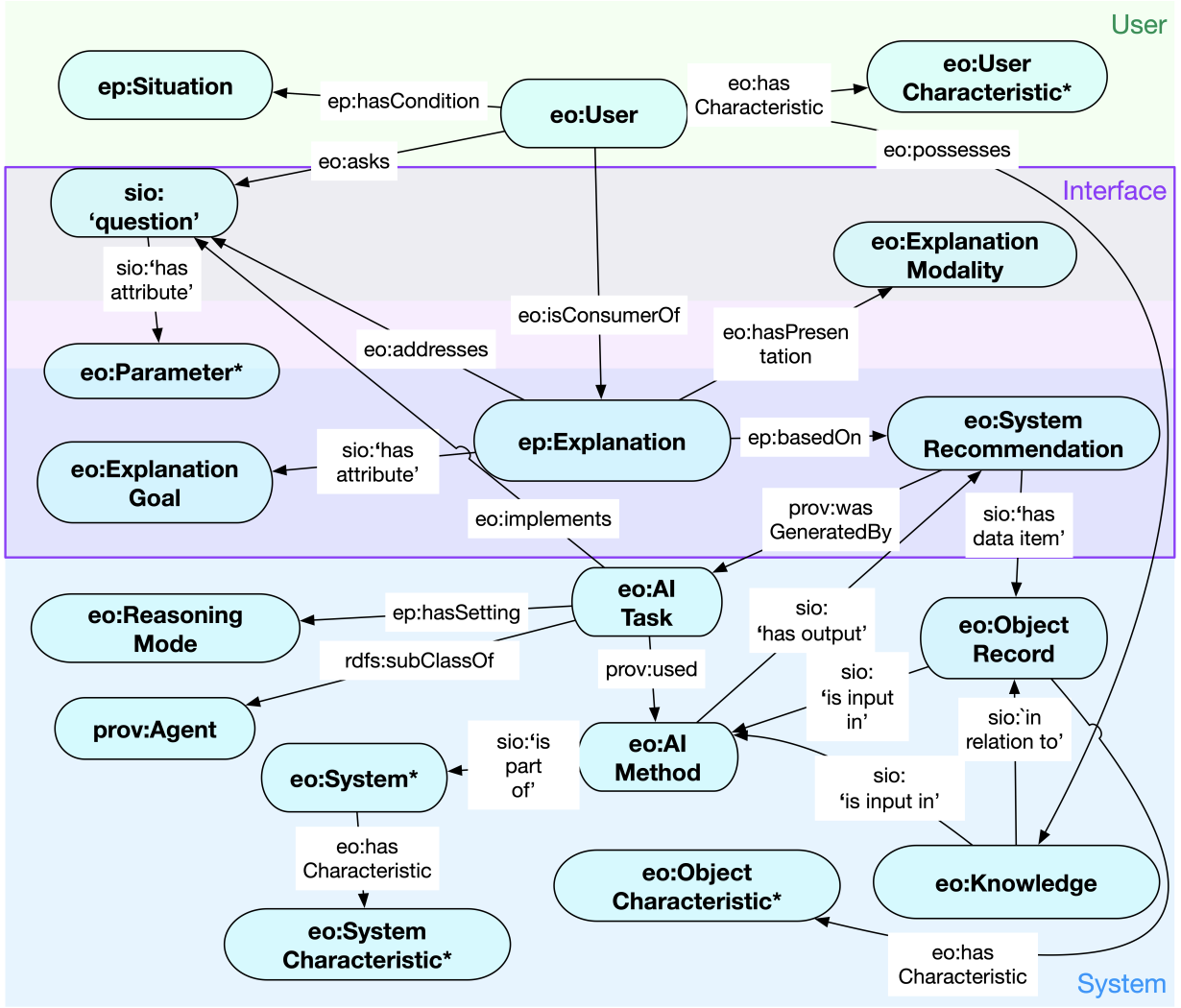}
    \caption{Explanation ontology overview with key classes separated into three overlapping attribute categories: user, interface, and system~\cite{Chari2023}.}
    \label{fig:explanation-onto}
\end{figure}

In a related study, Chari et al.~\cite{Chari2023} extended the ontology design pattern proposed by Tiddi et al.~\cite{Tiddi2015} to encompass explanations generated through computational processes. They developed an {\em explanation ontology} with the aim of facilitating user-centered explanations that enhance the explainability of model recommendations. This general-purpose model enables system designers to establish connections between explanations and the underlying data and knowledge. The explanation ontology incorporates attributes that form the foundation of explanations, such as {\em system recommendation} and {\em knowledge}, and models their interactions with the {\em questions}  being addressed. It broadly categorises the dependencies of explanations into user, system, and interface attributes (see~Figure~\ref{fig:explanation-onto}). 
User attributes capture concepts related to a user consuming an explanation, including the user's question, characteristics, and situation. System attributes encompass concepts associated with the AI system used to generate recommendations and produce explanations. Explanations are defined as a taxonomy of literature-derived explanation types, with refined definitions of nine specific types, namely {\em case-based}, {\em contextual},  {\em counterfactual}, {\em everyday}, {\em scientific}, {\em simulation-based}, {\em statistical}, and {\em trace-based} explanations. These explanation types serve different user-centric purposes and vary in their suitability for different user {\em situations}, {\em context} and {\em knowledge} levels. They are generated by various {\em AI Task} and {\em methods}, each with distinct informational requirements. Interface attributes capture the intersection between user and system attributes that can be directly interacted with on a user interface. The explanation ontology developed by Chari et al.~\cite{Chari2023} is publicly available at \url{https://tetherless-world.github.io/explanation-ontology}.

Overall, ontologies act as a lingua franca for representation and information exchange in explainable AI systems by providing a support shared, formal representation of domain knowledge. They facilitate transparent communication, promote collaboration between different stakeholders, and enhance the interpretability and reliability of AI systems' explanations. By using ontologies, XAI systems can bridge the gap between technical AI components and human understanding, making AI more transparent, accessible and, hence, trustworthy to end-users.

\section{Common-sense Reasoning}

The majority of existing approaches to XAI  rely on statistical analysis of black box models~\cite{IF-2023}. While this type of analysis has demonstrated its utility in gaining some insight into the internal workings of black box models, it generally lacks support for explainability based on  common-sense reasoning~\cite{Miller2017,WIREs-2020}. As a result, these approaches often fall short in providing explanations that closely align with human reasoning, thereby limiting their capacity to (in the best scenario) generating intelligible explanations. Conversely, there is widespread acceptance that symbolic knowledge can effectively facilitate common-sense reasoning. Therefore, it is reasonable to consider that explanation techniques can leverage ontologies to enhance model explainability and generate explanations that are more understandable to humans. 

In the context of XAI, several works attempt to fertilise explainability with ontologies. Seeliger et al.~\cite{Seeliger2019} surveyed what combinations of ontologies and knowledge graphs, and statistical models have been proposed to enhance model explainability, and what domains have been particularly important. 
The authors highlighted that quite a few approaches exist in supervised and unsupervised machine learning, whereas the integration of symbolic knowledge in reinforcement learning is almost overlooked. Most approaches in supervised learning seek to define a mapping between network inputs or neurons and ontology concepts which are then used in the explanations~\cite{Ribeiro_Leite_2021,AIJ-2021,Zachary2020}. 

\begin{figure*}[t!]
    \centering
    \subfloat[]
    {
        \includegraphics[width=.45\linewidth]{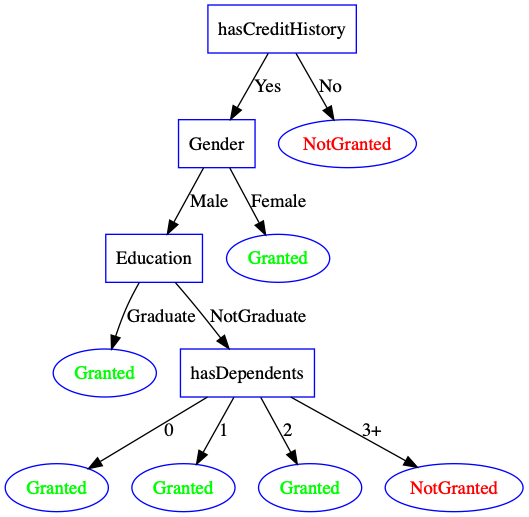}
    }
    ~
    \subfloat[]
    {
        \includegraphics[width=.45\linewidth]{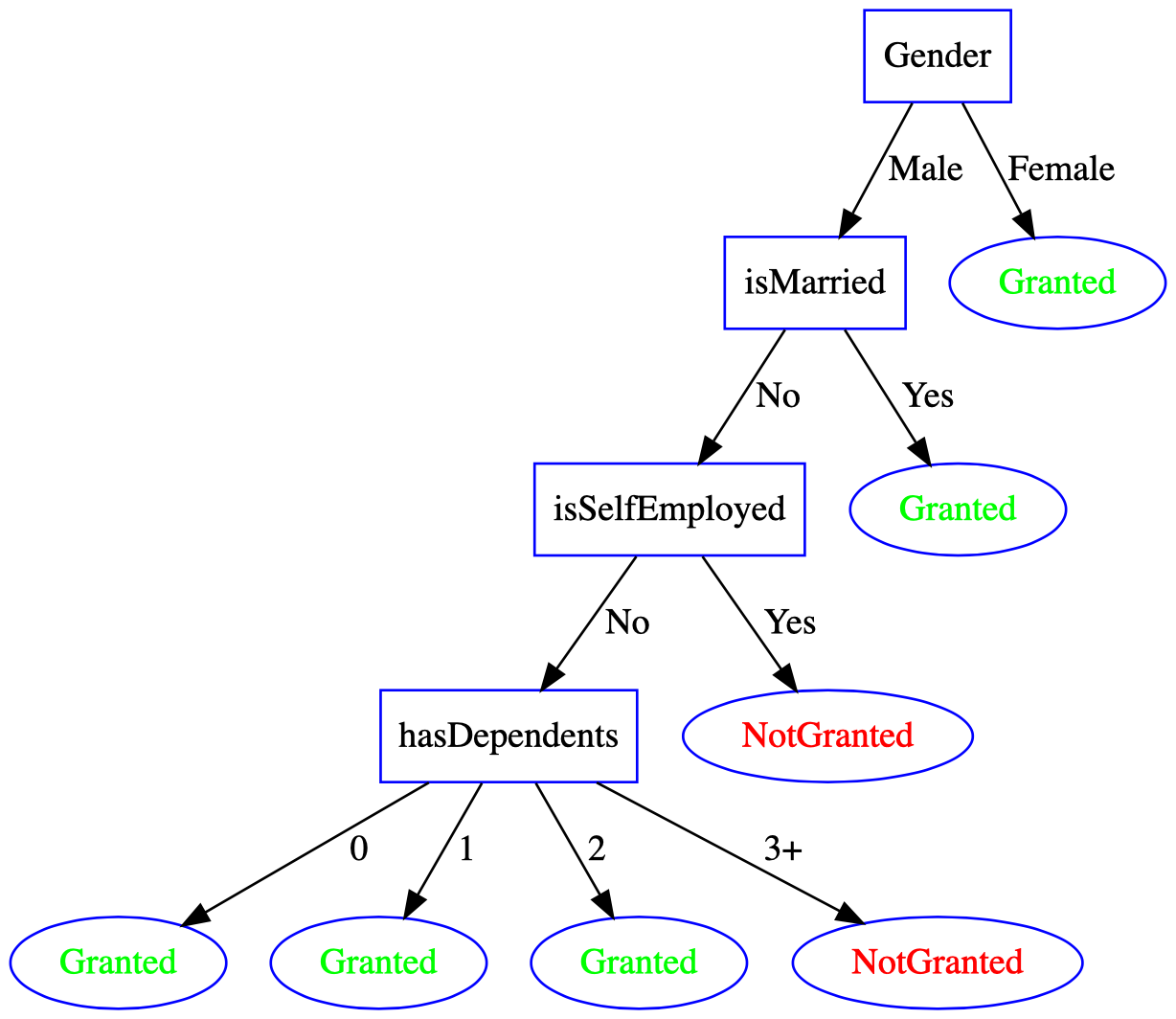}
    }
    \caption{Decision tree extracted without (a) and with (b) a domain ontology to explain the conditions to grant or refuse a loan~\cite{AIJ-2021}. It can be seen that the use of an ontology leads to different features appearing in the decision nodes. }
    \label{fig:trepan_reloaded}
\end{figure*}

In general, these methods rely on the presence of a domain ontology, which aids in generating symbolic justifications for the outputs of neural network models. Nevertheless, the way in which the ontology is integrated may differ among various approaches. 
In~\cite{Ribeiro_Leite_2021}, it is shown how the activations of the inner layers of a neural network wrt a given sample can be aligned with domain ontology concepts. To detect this alignment, multiple mapping networks (one network for each concept) are trained. Each network takes certain layer activations as input and produces the relevance probability for the corresponding concept as output. However, since the number of inner activations is typically substantial, and not all concepts can be extracted from each layer, this alignment procedure can be inefficient. 
In~\cite{Zachary2020}, it is assumed that training data contains labels that are (manually) mapped to concepts defined in a domain ontology. This semantic link is then exploited to provide explanations as justifications of the classification obtained. In~\cite{AIJ-2021}, a similar mapping is applied between input features and concepts in a domain ontology, where the ontology is used to guide the search for explanations. In particular, 
the authors proposed an algorithm that extracts decision trees as surrogate models of a black-box classifier and takes into account ontologies in the tree extraction  procedure. The algorithm learns decision trees whose decision nodes are associated with more general concepts defined in an ontology (Figure~\ref{fig:trepan_reloaded}). 
This has proven to enhance the human-understandability of decision trees~\cite{AIJ-2021}. 

In the context of unsupervised learning, Tiddi et al.~\cite{Dedalo2014} introduced a technique to explain clusters by traversing a knowledge graph in order to identify commonalities among them. The system generates potential explanations by utilising both the background knowledge and the given cluster, and it is independent of the specific clustering algorithm employed. 

One of the main challenges in all of these approaches lies in aligning the data utilised in statistical models with semantic knowledge. One possible solution is to create an ontology dedicated to each dataset and application, ensuring that the generated explanations are tailored to the specific problem. 
However, this approach can be prohibitive to application scenarios  with stringent time and scalability constraints. 
An alternative in these cases might be to systematically construct a suitable ontology for mapping. This involves mapping sets of features to existing general domain ontologies, such as MS Concept Graph~\cite{MSGraph12} or DBpedia~\cite{dbpedia-swj}. This process can be facilitated through ontology matching techniques \cite{GiunchigliaS03} and mapping methods~\cite{KalfoglouS03}. It is important to note that human supervision is required for this process, and manual fine-tuning of the mappings is necessary. Unfortunately, there is no definitive blueprint to follow in this regard. The inclusion of explicit knowledge is essential for any attempts at generating human-interpretable explanations. The choice between a domain-specific ontology or adapting a domain-independent (upper-level, foundational) one on an ad-hoc basis depends on the specific requirements of each application. 

There are ontology matching approaches in the literature that establish mappings between domain ontologies with support of foundational ontologies, as well as approaches that map domain ontologies to foundational ontologies. In general, mapping knowledge derived from statistical techniques to ontologies (as computational logical theories) allows for supporting a form of hybrid reasoning, in which symbolic automated reasoning can complement statistical ones and circumvent their limitations \cite{li2022combining}. However, in the particular case of mapping learned concepts and relations to a foundational ontology, we have the additional opportunity of grounding these knowledge elements in domain-independent axiomatized theories that describe common-sensical notions such as parthood, (existential, generic, historical) dependence, causality, temporal ordering of events, etc.~\cite{guizzardi2022ufo}. This can, in turn, support more refined and transparent explanations via the expansion of the consequences of these mappings enabled by logical reasoning. For example, by mapping an element E to an ontological notion of event in a event ontology \cite{botti2019representing}, one could infer a number of additional things about E (e.g., that it unfolds in time, that it has at least one object as participant, that it has a spatiotemporal extension, that its participants exist during the time the event is occurring, that it is a manifestation of some disposition, etc.).


\section{Knowledge Refinement and Complexity Management}

Abstraction and refinement are mechanisms that can be availed to represent knowledge in a more general or more specific manner. With abstraction one would hope to consider all is relevant and drop all the irrelevant details, with refinement one would hope to get more details. These mechanisms play a central role in human-reasoning. Humans use abstraction as a way to deal with the complexity of the world every day. Abstraction and refinement have many important applications, e.g., in natural language understanding, problem solving and planning, and reasoning by analogy. 

Different formalisation of abstraction and refinement were proposed in the literature.  
Keet~\cite{Keet07}  argued that most proposals for abstractions differ along three dimensions: language to which it is applied, methodology, and semantics of what one does when abstracting. A syntactic theory of abstraction, mainly based on proof-theoretic notions,  was proposed in \cite{GIUNCHIGLIA1992323}, whilst a semantic theory of abstractions based on viewing abstractions as model level mappings was proposed in~\cite{NayakL95}. These solutions  were mainly theoretical and not developed for or assessed on their potential for implementation, reusability and scalability.

Abstraction is tightly connected with  analogical reasoning. 
The structure mapping theory proposed by Gentner~\cite{GENTNER1983155} suggests that humans use analogical reasoning to map the structure of one domain    onto another, highlighting the shared relational information between objects. 
This process involves abstraction, as it can disregard specific object details in favor of identifying higher-level relational patterns that are relevant for making inferences and solving problems. 

In the context of Explainable AI, abstraction and analogical reasoning are essential for generating intelligible explanations to users. Abstraction allows an explanatory system to simplify complex decision-making processes by identifying higher-level patterns or by composing local explanations into (more general) global explanations~\cite{GLocalX}. On the other hand, analogical reasoning aids in aligning these patterns with familiar analogs from known domains. Consequently, users can grasp intricate decision-making processes in a more accessible and intelligible manner. 

Abstraction and analogical reasoning are closely linked to the idea that explanations are selective and can be seen as truthful approximations of reality~\cite{Miller2017}. One typical example is the scientific explanation of an atom through the analogy of a miniature solar system, where the nucleus plays the role of the sun and the electrons orbit around it like planets. Clearly, this analogy is a simplified representation of the atomic structure. In reality, atoms are much more complex, and the behavior of electrons is better described using quantum mechanics. However, by using the solar system analogy, lay users can visualise and understand the basic concept of how an atom structure is organised with a central nucleus and orbiting electrons. Furthermore, explanations often involve simplifications to make the information more understandable. These simplifications are necessary because the complete representation of reality can be  difficult to grasp. For instance, if asked to explain a prediction, a medical diagnosis AI system can offer an abstracted explanation that emphasizes the essential factors contributing to the diagnosis, rather than overwhelming the user with an exhaustive list of all the features and their values. 

\begin{figure*}[!t]
    \centering
    \includegraphics[width=\textwidth]{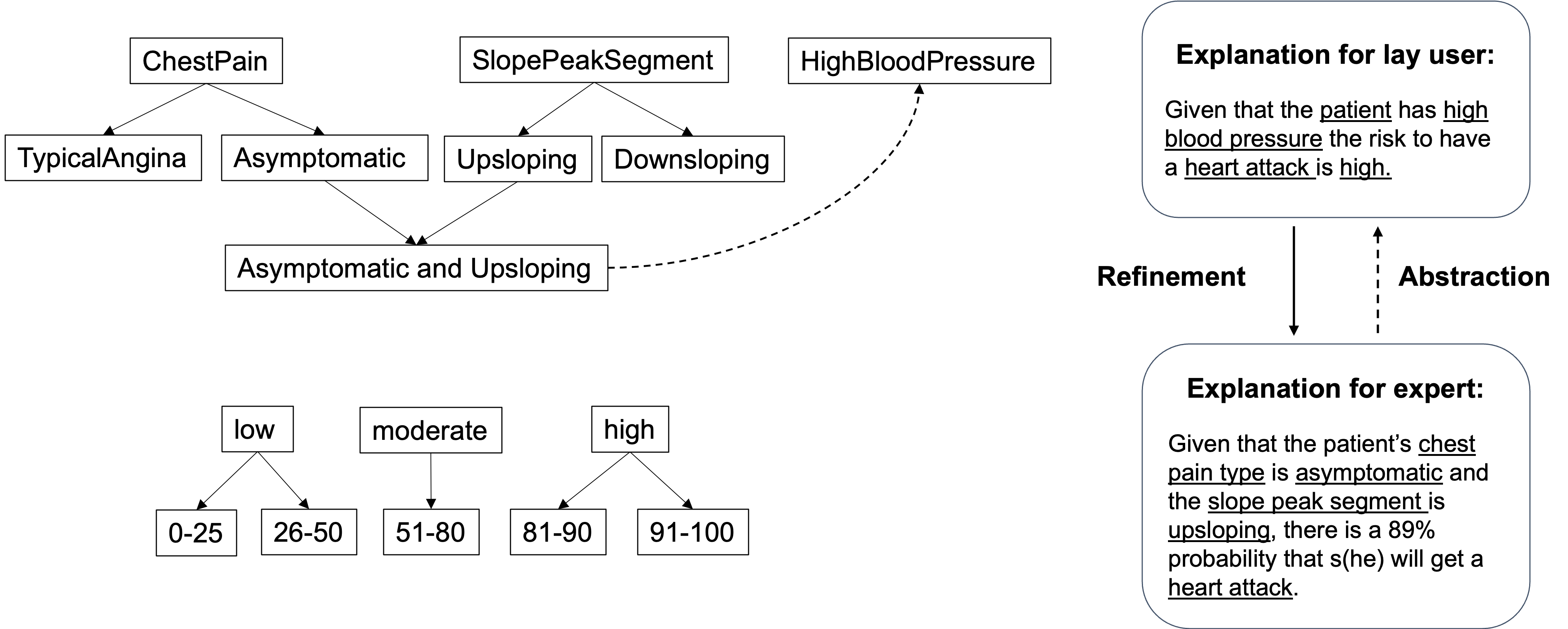}
    \caption{Refinement and abstraction applied to obtain explanations tailored to expert and lay users. Explanations are grounded to a domain ontology modeling concept definitions and relations between them. An explanation can be made more specific or more general by exploiting concept relationships defined in the ontology. In the explanation for expert users, the concepts associated with {\em asymptomatic chest pain} and {\em upsloping peak segment} can be replaced by the more general concept of {\em high blood pressure}. Similarly, the predicted  {\em probability value 89\%} of getting a {\em heart attack} can be replaced by its qualitative description {\em high}. In this way, it is is possible to obtain an explanation that abstracts from technical details and that can be more suitable for lay users. }
    \label{fig:abstraction_example}
\end{figure*}

To exemplify this idea further, let us imagine that the medical diagnosis AI system runs a predictive model about the risk of a patient of getting a heart attack, and that it needs to provide an explanation for each prediction. The recipient of the explanation can be a medical doctor, who is a domain expert, or a lay user (Figure~\ref{fig:abstraction_example}). 
On the one hand, the doctor would like to get a detailed   
explanation that provides insights about what attributes are used to make a diagnosis. On the other hand, a lay user might feel more comfortable in receiving a more abstract explanation that can still justify the diagnosis. Figure~\ref{fig:abstraction_example} exemplifies this idea. 
On the right side of the figure, two explanations are shown: one for an expert user and one for a lay user. Clearly, both of them explain the diagnosis but with a different level of details.\footnote{An explanation for a given instance can be obtained using a local explanation method such as LIME~\cite{Ribeiro2016} or ANCHOR~\cite{Ribeiro2018}. These methods explain the prediction of specific instances by identifying what attributes or features contribute more to the prediction. Once relevant attributes have been identified, a textual explanation can be generated by filling some explanation templates or by means of more advanced natural language processing techniques~\cite{rajani-etal-2019-explain}.}  
The explanations are grounded to a domain ontology, e.g., some features in the explanation are associated with concepts defined in the domain ontology. 
On the left side of the figure an excerpt of a simple ontology modeling the heart disease domain is represented. The ontology consists of simple concepts such as {\em ChestPain}, {\em Asymptomatic}, {\em TypicalAngina}, {\em SlopePeakSegment}, {\em Downsloping}, {\em Upsloping}, {\em HighBloodPressure}, a mapping between predicted probabilities of heart attacks to qualitative descriptions (e.g., a probability value in the range $[0,50]$ is mapped to {\em low}-risk), and relationships between them. Some concepts are more specific than others, for instance, {\em Asymptomatic} and {\em TypicalAngina} are types of {\em ChestPain}.  Simple concepts can be used to define  complex ones, for instance, the concept built by the conjunction of {\em Asymptomatic} and {\em Upsloping} defines {\em HighBloodPressure}. 
Once the explanations are grounded to the domain ontology, these can be made more specific or more abstract by exploiting the relationships defined in the ontology. For instance, the concepts related to chest pain type {\em Asymptomatic} and peak segment {\em Upsloping} can be replaced by {\em HighBloodPressure}. Similarly, the 89\% probability of getting a heart attack can be replaced by its qualitative description {\em high}-risk.  In this way, it is is possible to obtain an explanation that abstracts from technical details and that can be more suitable for lay users.  

Abstraction is just one of the possible techniques to address the goals of complexity management of complex information structures~\cite{OntoAbstraction2022}. Complexity Management, more generally, is essential for explanation, given that to explain one must focus on the explanation goals of an explanation-seeker~\cite{RomanenkoCG22}. In other works, according to the \textit{Pragmatic Approach to Explanation}~\cite{van1977pragmatics}, to explain is to answer the \textit{requests-for-explanation} (\textit{why-questions}) of an explanation-seeker. To do that, one has to dispense with information that does not contribute to that goal, i.e., to declutter the information structures and processes at hand. One can, for example, leverage the grounding of domain ontologies in foundational ones to automatically perform complexity management operations over large domain models. These operations include beside abstraction (ontology summarisation), modularisation and viewpoint extraction~\cite{Guizzardi2022,OntoAbstraction2022}.

Another important notion of refinement is {\em ontological unpacking}~\cite{guizzardi2023semantics}, a type of explanation (as well as type of knowledge refinement) that aims at revealing the ontological semantics underlying a given symbolic artifact. 
Ontological unpacking is a methodology that operationalises the notions of {\em truthmaking} and grounding in metaphysical explanation but also incorporating elements from the \textit{unificatory approach}~\cite{kitcher1981explanatory}, namely the use of \textit{foundational ontology design patterns} as an explanatory device. Design Patterns are the modeling analog of metaphors, i.e., they represent exactly the structure that is preserved across different (albeit analogous) situations. As discussed in~\cite{guizzardi2023semantics}, they can play a role analogous to the notion of a \textit{schematic structure/argument patterns} in the unificatory approach. In the latter, these patterns support explanation as \textit{theoretical reduction}, i.e., reducing (as in structural transferring~\cite{GENTNER1983155} or analogical reasoning) relevant aspects of a phenomenon to be explained to those of an understood phenomenon. In other words, theoretical reduction and unification using these patterns provide a mechanism for explanation employing analogy but also a mechanism for complexity management, since it puts focus on the elements that one needs to focus on, namely, exactly those constituting the pattern at hand. Ontological Unpacking has been shown to be effective in generating semantically transparent and unambiguous human-understandable explanations in complex domains~\cite{bernasconi2022semantic,garcia2023assessing}. It can be noted that ontology unpacking (as all refinement operations) goes in the inverse direction of abstraction~\cite{RomanenkoCG22}. Thus, ontology unpacking must be complemented by the aforementioned complexity management techniques. 

In summary, ontologies can support knowledge refinement and complexity management techniques, which in turn are key to creating intelligible explanations within explainable AI systems. By filtering out complex knowledge, refining explanations based on user needs and backgrounds, and aligning with human reasoning processes, an explanatory AI system can provide explanations that are not only transparent but also tailored to individual users, making AI decisions more interpretable and trustworthy.


\section{Discussion}\label{ses:discussion}

While ontologies may contribute to offer explanations within a domain of interest (see~e.g.,~\cite{AIJ-2021,Horne2019}), and, ontology usage should result in an understanding of the domain, this does not automatically mean that the ontology itself is self-explainable and easily understandable by humans~\cite{RomanenkoCG22}. The same holds for other symbolic artefacts that are  offered as explanations to numerical black boxes (e.g., knowledge graphs, decisions trees). For this reason, it can be argued that these symbolic artefacts also require their own explanation. The idea of ontological unpacking is relevant here as a means to identify and make explicit the \textit{truthmakers} of the propositions represented in that domain model, i.e., the entities in the world that make those propositions true, and from which the logico-linguistic constructions that constitute symbolic models are derived~\cite{guizzardi2023semantics}. This, in turn, can enhance the human understandability of explanations. 

A key notion related to explanations is that of  causality~\cite{causalityPearl2009}. Although not all explanations are causal explanations, causal explanations occupy a fundamental place in the scientific explanation literature \cite{kitcher1989explanatory}. In particular, knowing what relationship there is between input and output, or between input features can foster human-understandable explanations. However, causal explanations are largely lacking in the machine learning literature, with only few exceptions, e.g.,~\cite{CausalNN_ICML19}.  
Ontologies can capture causal rules once the knowledge over an application domain has been modelled. On the other hand, to the best of our knowledge, only a few works attempted to define and model causality within an ontology. In~\cite{CausalityBorgo2004}, the authors proposed different definitions of causality, and studied how constraints of different nature, namely structural, causality, and circumstantial,  intervene in shaping causal relations. However, as the authors claim, the approach is incomplete and further extensions are needed. \cite{botti2019representing} puts forth a formal theory of causation. In this theory, we have the explicit manifestation of dispositions, which are activated by the obtaining of certain situations (roughly individual state of affairs), and which are manifested via the occurrence of events. In other words, events cause each other via a continuous mechanism of events bringing about situations, that activate dispositions, that in turn are manifested as other events, and so on.

Another ontology-based approach to explanation, which is connected to value-based justification (in ethics) is the one discussed in~\cite{guizzardi2023ontology}. There, explicability is related to the reconstruction of decision-making processes, which in turn are grounded on preference relations, which in turn are grounded on value-assessments. The whole approach is grounded on an ontological analysis of ethical dimensions, and, ultimately, on an ontological analysis of the notions of value, risk, autonomy and delegation.

Finally, we highlight the open challenge of empirically evaluating the human-understandability and effectiveness of explanations. Rudin et al.~\cite{rudin22} pointed out evaluation of explanations as a major challenge to be faced in the context of XAI. On the one hand, one would like to quantify to what extent user can  understand and use an explanation. A few approaches proposed  quantitative metrics and protocols but any consensus has been reached by the community yet.  Hoffman et al.~\cite{HoffmanMMKC18} introduced key concepts for measuring the quality of an XAI system derived from the integration of extensive research literature and psychometric assessments. Vilone and Longo \cite{VILONE202189} aggregated all the scientific studies that classify theories and notions related to the concept of explainability  as well as the evaluation approaches for XAI methods via a hierarchical system. Cromik and Schuessler~\cite{Chromik2020ATF} proposed some tasks to be used in human-evaluation of explanations. An analysis of these work reveals that evaluating the goodness and effectiveness of automated explanations is a prerequisite for ensuring calibrated trust in AI~\cite{Nauta23}. On the other hand, there is the problem of how to measure the causal understanding of an explanation (the causability of an explanation)~\cite{Holzinger2019b-causability}. 
Whilst this is always possible for explanations of human statements (as the explanation is per-se related to a human model), measuring the causal understanding of an explanation of a machine statement has to be based on a causal model, which is not the case for most machine learning algorithms~\cite{Holzinger2020CausabilityScale}.


\section{Summary and Conclusion}\label{sec:conclusion}

In this paper, we discussed the role of ontologies and knowledge in Explainable AI from three perspectives: reference modeling, common-sense reasoning, and knowledge refinement and complexity management. 

The role played by ontologies within these perspectives can be summarised as follows. Firstly, 
ontologies provide formal reference consensual models for designing explainable systems and generating human-understandable explanations. Ontologies provide a common lingua for defining explanations, promoting interoperability, and reusability of explanations across various domains. Secondly, ontologies enable the creation of explanations with linked semantics. This can, in turn, support more refined and transparent explanations via knowledge expansion enabled by logical reasoning. Thus, integrating ontologies with current explanation techniques allows for supporting a form of hybrid reasoning, enhancing the human-understandability of explanations. Finally, ontologies offer the ability to abstract and refine knowledge, which serves as the foundation for human reasoning. Knowledge refinement and complexity management is  essential to craft personalised explanations that are human-centric and tailored to different user profiles.

Given the above, ontologies can play a crucial role in the realm of Explainable AI. Nonetheless, a number of challenges still need to be addressed, namely the integration of foundational and domain ontologies in current explainability approaches, the adoption of complexity management techniques to ensure ontologies as explanations are easily manageable and comprehensible for users, and the evaluation of their human-understandability.

\end{document}